\documentclass{article} 
\usepackage{aiarxiv}
\usepackage{times}
\usepackage{url}
\usepackage{amssymb}
\usepackage{amsmath}
\usepackage{relsize}

\usepackage{xcolor}

\usepackage{booktabs}
\usepackage{multirow}
\usepackage{siunitx}

\usepackage{soul}

\definecolor{navy}{rgb}{.25,.25,.75}
\definecolor{ruby}{rgb}{.70,.20,.20}
\definecolor{magenta}{rgb}{1.0,.0,1.0}

\newcommand{\surl}[1]
{
	\urlstyle{same}\url{#1}
}
\newcommand{\etal}{et al.}

\usepackage{accents}

\usepackage{mathtools}

\newcommand{\eat}[1]{}
\newcommand{\rev}[1]{#1}
\newcommand{\verify}[1]{#1}
\newcommand{\rvx}[1]{#1}
\newcommand{\rva}[1]{{{#1}}}

\newcommand{\bfheading}[1]{\noindent{\bf{#1}}}

\newcommand{\secmoveup}{\vspace{-0.mm}} 

\newcommand{\textmoveup}{\vspace{-0.mm}} 
\newcommand{\eqmoveup}{\vspace{-0.mm}}                
\newcommand{\captionmoveup}{\eqmoveup\vspace{-1.mm}}  

\newcommand{\0}{\mathbf{0}}
\newcommand{\x}{\mathbf{x}}
\newcommand{\y}{\mathbf{y}}

\newcommand{\h}{\mathbf{h}}
\newcommand{\vc}{\mathbf{c}}
\newcommand{\vi}{\mathbf{i}}
\newcommand{\vo}{\mathbf{o}}
\newcommand{\f}{\mathbf{f}}
\newcommand{\g}{\mathbf{g}}
\newcommand{\Ey}{\mathbf{Ey}}
\newcommand{\E}{\mathbf{E}}

\newcommand{\W}{\mathbf{W}}
\newcommand{\T}{\mathbf{T}}
\newcommand{\Y}{\mathbf{Y}}
\newcommand{\bbR}{{\mathbb R}}
\newcommand{\fracp}[2]{\frac{\partial #1}{\partial #2}}

\title{SentiCap: Generating Image Descriptions with Sentiments}

\author{Alexander Mathews$^*$,~Lexing Xie$^{*\dag}$,~Xuming He$^{\dag *}$\\
$^*$The Australian National University, $^\dag$NICTA\\
alex.mathews@anu.edu.au, lexing.xie@anu.edu.au, xuming.he@nicta.com.au}

\begin{document}

\maketitle
\begin{abstract}\textmoveup
The recent progress on image recognition and language modeling 
is making automatic description of image content a reality. 
However, stylized, non-factual aspects of 
the written description are missing from the current systems. 
One such style is descriptions with emotions, 
which is commonplace in everyday communication, and 
influences decision-making and interpersonal relationships.  
We design a system to describe an image with emotions, 
and present a model that automatically generates captions with 
positive or negative sentiments. 
We propose a novel switching recurrent neural network with word-level regularization, 
which is able to produce emotional image captions using 
only 2000+ training sentences containing sentiments. 
We evaluate the captions with different 
automatic and crowd-sourcing metrics. 
Our model compares favourably in common quality metrics for image captioning. 
In 84.6\% of cases the generated positive captions were judged as being at 
least as descriptive as the factual captions. Of these positive captions 88\% were 
confirmed by the crowd-sourced workers as having the appropriate sentiment.
\end{abstract}

\secmoveup
\section{Introduction}
\label{sec:intro}

\begin{figure}[t]
  \centering
  \includegraphics[width=.4\textwidth]{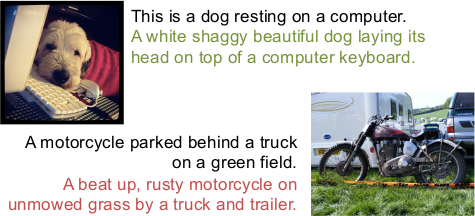}
  \caption{Example images with neural, positive (\textcolor{green}{green}) and negative (\textcolor{red}{red}) captions, by crowd workers in MSCOCO dataset~\protect\cite{chen2015microsoft} and this work (Section~\protect\ref{sec:mturk}). }
  \label{fig:intro}
\end{figure}

Automatically describing an image 
by generating a coherent sentence unifies two
core challenges in artificial intelligence -- 
vision and language. Despite being a difficult problem, 
the research community has recently made headway into this area, 
thanks to large labeled datasets,
and progresses in learning expressive  
neural network models. 
In addition to composing a factual description about the objects, 
scene, and their interactions in an image, 
there are richer variations in language, 
often referred to as styles~\cite{crystal1969investigating}. 
Take emotion, for example, it is such a common phenomena in 
our day-to-day communications that over half of text 
accompanying online pictures contains an emoji (a graphical alphabet for emotions)~\cite{instagram2015emoji}.
How well emotions are expressed and understood 
\rvx{influences} decision-making~\cite{lerner2015emotion} -- from the mundane (e.g., making a restaurant menu appealing) to major (e.g., choosing a political leader in elections).
Recognizing \rvx{sentiment and opinions} from written communications has been an active research topic for the past decade~\cite{pang2008opinion,socherrecursive}, the synthesis of text with sentiment that is relevant to a given image is still an open problem. 
In Figure~\ref{fig:intro}, \rvx{each} image is described with a factual caption, and with positive or negative emotion\rvx{, respectively}. One may argue that the descriptions with sentiments are more likely to pique interest 
about the subject being pictured (the dog and the motocycle), or about their background settings (interaction with the dog at home, or how the motocycle came about). 

In this paper, we describe a method, called SentiCap, to generate image captions with sentiments. We build upon the CNN+RNN (Convolution Neural Network + Recurrent Neural Network) recipe that has seen many recent successes~\cite{donahue2015long,Karpathy2015CVPR,mao2014deep,vinyals2015show,xu2015show}.
In particular, we propose a switching Recurrent Neural Network (RNN) model to represent sentiments. 
This model consists of two parallel RNNs -- 
one represents a general background language model; another specialises in descriptions with sentiments. 
We design a novel word-level regularizer\eat{ for the sentence likelihood}, 
so as to emphasize the sentiment words during training \rvx{and to optimally combine the two RNN streams} (Section~\ref{sec:model}). We have gathered a new dataset of several thousand captions with positive and negative sentiments by re-writing factual descriptions (Section~\ref{sec:mturk}). 
\rvx{Trained on 2000+ sentimental captions and 413K neutral captions, our switching RNN out-performs a range of heuristic and learned baselines 
in the number of emotional captions generated, and in a \rva{variety} of subjective and human evaluation metrics.
In particular SentiCap has the highest number of success in placing at least one sentiment word into the caption, 
88\% positive (or 72\% negative) captions are perceived by crowd workers as more positive (or negative) than the factual caption, with a similar descriptiveness rating. }

\eat{demonstrates a notable improvement from the original, fine-tuned, and domain-adapted models in automatic evaluation as measured by {\sc Bleu}, {\sc Meteor}, {\sc Cider}. }
\eat{which show that our apporach generates captions with sentiment more frequently than the baseline methods, and is at least as descriptive as the purely facutal captions in a large proportion of cases.}
\eat{ we first found that the vast majority of images can be described with either positive or negative sentiment, making the image captioning problem not purely objective.  We also observed that the descriptiveness of the positive captions is generally on par with those of the factual captions, and that our model is able to generate more emotional captions than the alternatives. }

\secmoveup
\section{Related Work}

\eat{The number of visual objects and scenes that can be reliably recognized has leaped from dozens to thousands in the past few years, largely due to the advances in training Convolutional Neural Networks (CNN)~\cite{Simonyan2015,Szegedy_2015_CVPR}. }

\rvx{Recent} advances in visual recognition have made ``an image is a thousand words'' much closer to reality, \rvx{largely due to the advances in Convolutional Neural Networks (CNN)~\cite{Simonyan2015,Szegedy_2015_CVPR}}. 
\rvx{A related topic also advancing rapidly is image captioning, where most early systems were}
\rev{based on similarity retrieval using objects and attributes~\cite{farhadi2010every,kulkarni2011baby,hodosh2013framing,Gupta2012a}, and assembling sentence fragments \eat{from relationships }
such as object-action-scene~\cite{farhadi2010every}, subject-verb-object~\cite{rohrbach2013translating}, object-attribute-prepositions~\cite{kulkarni2011baby} or global image properties such as scene and lighting~\cite{Nwogu2011}.} 
\rvx{Recent systems model richer language structure, such as formulating a integer linear program to map visual elements to the parse tree of a sentence~\cite{kuznetsova2014treetalk}, or embedding ~\cite{Xu2015}  video and compositional semantics into a joint space.}
\eat{, and uses an integer linear program to find good tree configurations. 
Xu~\etal~\shortor jointly embedding of video and compositional semantics. }
\eat{Some recent works have also looked at the }

\eat{The construction of }Word-level language models \rvx{such as RNNs~\cite{mikolov2011strategies,sutskever2011generating} and maximum-entropy (max-ent) language models~\cite{mikolov2011strategies} have improved with the aid of} significantly larger datasets and more computing power.
Several research teams independently proposed image captioning systems that combine CNN-based image representation and \rvx{such} language models. Fang \etal~\shortcite{fang2015CVPR} used a cascade of word detectors from images and a \rvx{max-ent} model. The Show and Tell~\cite{vinyals2015show} system used an RNN as the language model, seeded by CNN image features. Xu \etal~\shortcite{xu2015show} estimated spatial attention as a latent variable, to make the Show and Tell system aware of local image information. Karpathy and Li\rvx{~\shortcite{Karpathy2015CVPR}} used an \rvx{RNN to generate a sentence from the alignment between objects and words}.
\rev{Other work has}\eat{Some authors have } \rev{employed multi-layer RNNs~\cite{Chen_2015_CVPR,donahue2015long}} \rvx{for image captioning.}
\rev{Most} RNN-based multimodal language models\eat{~\cite{donahue2015long,vinyals2015show,xu2015show}} \rev{use} the Long Short Term Memory (LSTM) unit that preserves long-term information \rvx{and prevents}\eat{while preventing} overfitting~\cite{hochreiter1997long}. We adopt one of the competitive systems~\cite{vinyals2015show} -- CNN+RNN with LSTM units as our basic multimodal sentence generation engine, due to its simplicity and computational efficiency. 

\eat{and then used this alignment to train an RNN for generating sentences. }
\eat{to model the relationship between words and images.}

\eat{The research community have not only modeled what are in the images, but also }
\rvx{Researchers have modeled} how \rvx{an} image is presented, and what kind of response it is likely to elicit from \eat{human }viewers, such as analyzing the aesthetics and emotion in images~\cite{murray2012ava,joshi2011aesthetics}. 
More recently, the Visual SentiBank~\cite{borth2013sentibank} system constructed a catalogue of Adjective-Noun-Pairs (ANPs) that are frequently used to describe online images.
We build upon Visual SentiBank to construct sentiment vocabulary\eat{ for describing images}, but to the best of our knowledge, no existing work tries to compose image descriptions with desired sentiments. \rev{Identifying sentiment in text is an active area of research\rvx{~\cite{pang2008opinion,socherrecursive}}. Several \rvx{teams}~\cite{Nakagawa2010,Mcdonald2011} \eat{have modelled sentiment with hidden variables represent the sentiment expressed in sentences. }
\rvx{designed sentence models with latent variables representing the sentiment.}
\eat{We differ from this work as we are both }\rvx{Our work focuses on}
generating sentences and not explicitly modelling sentiment using hidden variables.}

\secmoveup
\section{Describing an Image with Sentiments}
\label{sec:model}

Given an image $I$ and its $D_x$-dimensional visual feature $\x\in\bbR^{D_x}$, our goal is to generate a sequence of words (i.e. a caption) $\Y=\{\y_1,\cdots,\y_T\}$ to describe the image with a specific style, such as expressing sentiment. Here $\y_t\in\{0,1\}^V$ is 1-of-V encoded indicator vector for the $t^{th}$ word; $V$ is the size of the vocabulary; and $T$ is the length of the caption. 

We assume \rva{that} sentence generation involves two underlying mechanisms, one of which focuses on the factual description of the image while the other describes the image content with sentiments. 
\rev{We formulate such caption generation process using a switching multi-modal language model, which sequentially generates words in a sentence.}
\rev{Formally}, we introduce a binary \rev{sentiment} variable $s_t\in\{0,1\}$ for every word $\y_t$ to indicate which mechanism is used. 
At each time step $t$, our model produces the probability of $\y_t$ and the current sentiment variable $s_t$ given the image feature $\x$ and the previous words $\y_{1:t-1}$, denoted by $p(\y_t,s_t|\x,\mathbf{y}_{1:t-1})$. We generate the word probability by marginalizing \rev{over} the sentiment variable $s_t$:
\begin{equation}
\resizebox{0.25\columnwidth}{!}{$
p(\y_t|\x,\y_{1:t-1})$}=\sum_{s_t}\resizebox{0.525\columnwidth}{!}{$p(\y_t|s_t,\x,\y_{1:t-1})p(s_t|\x,\y_{1:t-1})
$}
\label{eq:bayesrule_s}
\eqmoveup
\end{equation}
Here $p(\y_t|s_t,\cdot)$ is the caption model conditioned on the sentiment variable and \rev{$p(s_t|\cdot)$ is the probability of the word sentiment}. The rest of this section will introduce these components \rev{and model learning} in detail.

\secmoveup
\subsection{Switching RNNs for Sentiment Captions}
\label{ssec:sentimodel}

We adopt a joint CNN+RNN architecture~\cite{vinyals2015show} in the conditional caption model. Our full model combines two CNN+RNNs running in parallel: 
one capturing the \rev{factual word generation (referred to as the background language model)}, 
the other specializing in words with sentiment. \rev{The full model is a switching RNN, in which the variable $s_t$ functions as a switching gate.} This model design aims to learn sentiments well, despite data sparsity --\eat{- e.g.,} using only a small dataset \rvx{of}\eat{on} image description with sentiments (Section~\ref{sec:mturk}), with the help from millions of image-sentence pairs that factually describe pictures~\cite{chen2015microsoft}.

\eat{We propose a pair of RNNs for image description with sentiments, which 
consists of two RNNs running in parallel. }
\eat{Each of the two RNNs is denoted with $k=0 \text{ or } 1$. }
\eat{Given \verify{the sentiment switch variable $s_t=k$} indicating which RNN stream it is,}

Each RNN stream \eat{is built with}\rvx{consists of} a series of LSTM units. Formally, we denote the $D$-dimensional hidden state of an LSTM as $\h_t \in \bbR^D$, its memory cell as $\vc_t\in \bbR^D$, the input, output, forget gates as $\vi_t,~\vo_t,~\f_t \in \bbR^D$, \rev{respectively.}
\rev{Let $k$ indicate which RNN stream it is,} the LSTM can be implemented as: 
\begin{align}
&\begin{pmatrix} \vi_t^k \\ \f_t^k \\ \vo_t^k \\ \g_t^k \end{pmatrix}  = 
\begin{pmatrix} \sigma \\ \sigma \\ \sigma \\\tanh \end{pmatrix} \mathlarger{\T}^k\eat{_{4D, 2D}}
\begin{pmatrix} \E^k\y_{t-1} \\ \h^k_{t-1} \end{pmatrix}
\label{eq:lstm}\\
&\vc_t^k = \f_t^k \odot \vc_{t-1}^k + \vi_t^k \odot \g_t^k, \quad
\h_t^k = \vo_t^k \odot {\vc_t^k}. \nonumber
\eqmoveup
\end{align}
Here $\sigma(\chi)$ is the sigmoid function $1/(1+e^{-\chi})$; $tanh$ is the hyperbolic tangent function;
\rvx{$\T^k\in\bbR^{4D\times2D}$} is a set of learned weights; 
$\g^k_t\in\bbR^D$ is the input to the memory cell;
\rvx{$\E^k \in \bbR^{D\times V}$} is a learned embedding matrix  \rvx{in model $k$}, and $\E^k\y_t$ is the embedding vector of the word $\y_t$.

To incorporate image information, we use an image representation $\hat\x=\W_x\x$ as the word embedding $\Ey_0$ when $t=1$, where $\x$ is a high-dimensional image feature extracted from a convolutional neural network~\cite{Simonyan2015}, and $\W_x$ is a learned embedding matrix.    
Note that the LSTM hidden state $\h_t^k$ summarizes $\y_{1:t-1}$ and $\x$.
The conditional probability of the output caption words depends on the hidden state of the corresponding LSTM,    
\begin{align}
p(\y_t | s_t=k, \x, \y_{1:t-1}) &\propto \exp(\W_y^{k} \h_t^k) 
\eqmoveup
\end{align} 
where $\W_y^k \in \bbR^{D\times V}$ is a set of learned output weights. 

\rvx{The} sentiment switching model\eat{The sentiment model }
generates the probability of switching \rvx{between the two RNN streams} at each time $t$, 
\rvx{with}\eat{We design} a single layer network taking the hidden states of both RNNs as input:
\begin{align}
p(s_t=1 | \x, \y_{1:t-1}) &= \sigma(\W_s[\h^0_t;\h^1_t])\label{eq:switch}
\eqmoveup
\end{align} 
where $\W_s$ is the weight matrix for the hidden states. 

\begin{figure}[t]
  \centering
  \includegraphics[width=.35\textwidth]{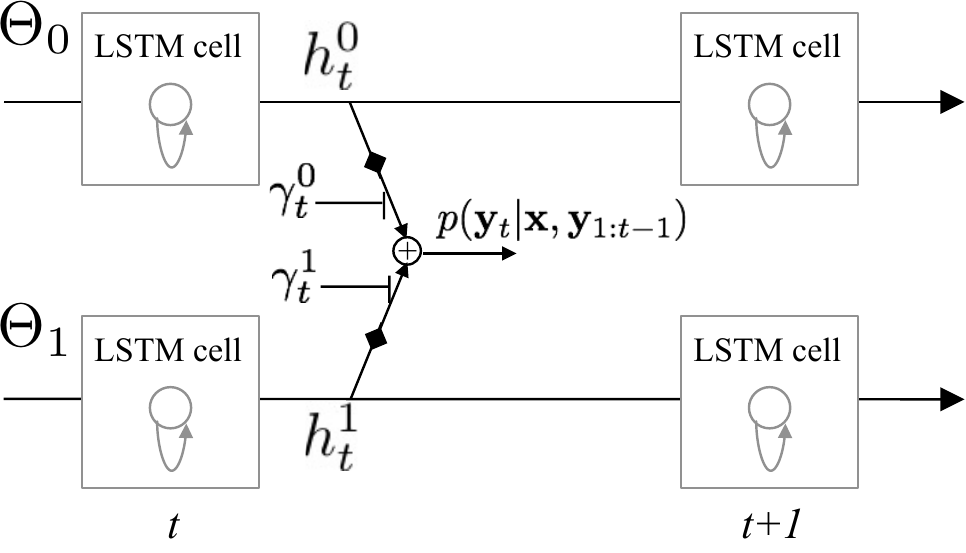}
  \caption{Illustration of the switching RNN model for captions with sentiment. Lines with diamonds denote projections with learned weights. LSTM cells are described in Eq~\ref{eq:lstm}. $\gamma_t^0$ and $\gamma_t^1$ are probabilities of sentiment switch defined in Eq~(\protect\ref{eq:switch}) and act as gating functions for the two streams. 
  }
  \label{fig:rnn}\captionmoveup
  \vspace{-4.0mm}
\end{figure}

\rvx{An illustration of this sentiment switching model is in Figure~\ref{fig:rnn}.} 
In summary, the parameter set for each RNN ($k=\{0,1\}$) is \rvx{$\Theta^k=\{\T^k, \W_y^k, \E^k, \W_x^k\}$}, and that of the switching RNN is \rvx{$\Theta=\Theta^0\cup\Theta^1\cup \W_s$}. 
We have tried including $\x$ for learning $p(s_t|\x, \y_{1:t-1})$ but found no benefit. 

\secmoveup
\subsection{Learning the Switching RNN Model}
\textmoveup

\rev{One of the key challenges is to design a learning scheme for $p(s_t | \x, \y_{1:t-1})$ and two CNN+RNN components. We take a two-stage learning approach to estimate the parameters $\Theta$ in our switching RNN model based on a large dataset with factual captions and a small set with sentiment captions.}  

\noindent\textbf{Learning a background multi-modal RNN.} We first train a CNN+RNN with a large dataset of image and caption pairs, denoted as $\mathcal{D}^0=\{(\x_0^i,\y_0^i)\}_{i=1}^N$. 
$\Theta^0$ are learned by minimizing the negative log-likelihood of the caption words given images, 
\rev{
\begin{equation}\label{eqn:baselearn}
\resizebox{0.26\columnwidth}{!}{$	
	L^0(\Theta^0,\mathcal{D}^0)
= -$}\sum_i\sum_t \resizebox{0.5\columnwidth}{!}{$\log p(\y^i_{0,t}|s_t=0,\x^i_0,\y^i_{0,1:t-1} ).$}
\eqmoveup
\end{equation}
}

\noindent\textbf{Learning from captions with sentiments.} \rev{Based on the pre-trained CNN+RNN in Eq~\eqref{eqn:baselearn}, we then learn the switching RNN using a small image caption dataset with a specific sentiment polarity, denoted as $\mathcal{D}=\{(\x^i,\y^i,\eta^i)\}_{i=1}^M$, $M\ll N$. Here $\eta_t^i \in [0, 1]$ is the sentiment strength of the $t^{th}$ word in the $i$-th training sentence,} \rvx{being either positive or negative as specified in the training data.}
\eat{, with its absolute value $\bar\eta_t^i = |\eta_t^i| \in [0, 1]$ denoting sentiment strength, and $sign(\eta_t^i)$ denoting sentiment polarity -- positive or negative.}

\eat{This is achieved in two steps.}
\eat{\rev{The first \verify{step} is to initialize} $\Theta^1$ \rev{by} regularized fine-tuning on dataset $\cal D$. \rvx{Such fine-tuning} finds a trade-off between \rvx{data} likelihood and the L2 difference between the current and base RNN, \rvx{and is one of the most competitive approaches in domain transfer~\cite{schweikert2008empirical}}.
	\verify{
		\begin{align}
		\min_{\Theta^1} \resizebox{0.15\columnwidth}{!}{$L'(\Theta,\mathcal{D})$}
		&= -\sum_i\sum_t \log p(\y^i_{t}|\x^i,\y^i_{1:t-1} ) + R(\Theta),\nonumber\\
		R(\Theta)&=\frac{\lambda_\theta}{2}\|\Theta^1 - \Theta^0\|^2 
		\eqmoveup
		\end{align}
	}
} 

\rev{We design a new training objective function to use word-level sentiment information for learning $\Theta^1$ and the switching weights $\W_s$, while keeping the pre-learned $\Theta^0$ fixed.
For clarity, we denote the sentiment probability as:
\begin{align}
\gamma^{0}_t = p(s_t = 0|\x,\y_{1:t-1}), \quad
\gamma^{1}_t = 1-\gamma^{0}_t;  \label{eq:switch}
\eqmoveup
\end{align}
and the log likelihood of generating a new word $\y_t$ given image and word histories $(\x,\y_{1:t-1})$ as $L_t(\Theta,\x,\y)$, which can be written as  
(cf. Eq~\eqref{eq:bayesrule_s}), 
\begin{align}
L_t&(\Theta,\x,\y)=  \log p(\y_t | \x, \y_{1:t-1})= \\ 
&\log [\gamma^0_t p(\y_t|s_t=0,\x,\y_{-t} ) + 
 \gamma^1_t p(\y_t|s_t=1,\x,\y_{-t} )].\nonumber   
\end{align}
The overall learning objective function for incorporating word sentiment is a 
combination of a weighted log likelihood and the cross-entropy between $\gamma_t$ and \rev{$\eta_t$},
\begin{align}
{\cal L}(\Theta,\mathcal{D}) &= -\sum_i\sum_t (1+ \lambda_\eta \eta^i_t ) 
[L_t(\Theta,\x^i,\y^i)  \label{eq:giantr}
\\ &+ \lambda_\gamma (\eta_t^i \log\gamma^{1,i}_t + (1-\eta_t^i) \log\gamma^{0,i}_t)  ]  + R(\Theta),\nonumber\\
R(\Theta)=&\frac{\lambda_\theta}{2}\|\Theta^1 - \Theta^0\|^2 
\label{eq:modreg}
\end{align}
where $\lambda_\eta$ and $\lambda_\gamma$ are weight parameters, and $R(\Theta)$ is the regularization term with weight parameter $\lambda_\theta$. Intuitively,} when $\eta_t > 0$, i.e. the training sentence encounters \rvx{a sentiment word}, the likelihood weighting factor $\lambda_\eta \eta^i_t$ \rev{increases the importance of $L_t$ in the overall likelihood; at the same time, the cross-entropy term $\lambda_\gamma (\eta_t^i \log\gamma^{1,i}_t + (1-\eta_t^i) \log\gamma^{0,i}_t) $ encourage switching variable $\gamma^1_t$ to be $>0$, emphasizing the new model. 
The regularized training finds a trade-off between the \rvx{data} likelihood and L2 difference between the current and base RNN, \rvx{and is one of the most competitive approaches in domain transfer~\cite{schweikert2008empirical}}.}

\noindent\bfheading{Settings for model learning.} 
We use stochastic gradient descent with backpropagation on mini-batches to optimize the RNNs. We apply dropout to the input of each step, which is either the image embedding $\hat\x$ \rvx{for $t=1$} or the word embedding $\E^k\y_{t-1}$ and the hidden output $\h^k_{t-1}$ from time $t-1$, for both the background and sentiment streams $k=0,1$.

We learn models for positive and negative sentiments separately, due to the observation that either sentiment could be valid for the majority of images (Section~\ref{sec:mturk}). 
\rev{We initialize $\Theta^1$ as $\Theta^0$ and use 
the following gradient of to minimize ${\cal L}(\Theta,\mathcal{D})$  with 
respect to $\Theta^1$ and $\W_s$, holding $\Theta^0$ fixed.
\begin{align}
\fracp{\cal L}{\Theta} = &-\sum_i\sum_t (1+ \lambda_\eta \eta^i_t ) [\fracp{L_t}{\Theta} \nonumber\\&+ \lambda_\gamma ( \frac{\eta_t^i}{\gamma^{1,i}_t}\fracp{\gamma^{1,i}_t}{\Theta} + \frac{1-\eta_t^i}{\gamma^{0,i}_t} \fracp{\gamma^{0,i}_t}{\Theta}) ] +  \fracp{R(\Theta)}{\Theta}
\eqmoveup
\end{align}
}Here $\fracp{L_t}{\Theta},\fracp{\gamma^{0,i}_t}{\Theta},\text{ and }\fracp{\gamma^{1,i}_t}{\Theta}$ are computed through differentiating across Equations~(\ref{eq:bayesrule_s})--(\ref{eq:switch}). 
During training, we set $\eta_t=1$ when word $\y_t$ is part of an ANP with the target sentiment polarity, otherwise $\eta_t=0$. We also include a default L2-norm regularization for neural network tuning $|\Theta|^2$ with a small weight ($10^{-8}$). \eat{We found that regularized fine-tuning in Equation~\ref{eq:model1r} outperforms fine-tuning, and that regularized fine-tuning shall be carried out jointly with word-level sentiment switches, i.e. use ${\cal L}(\Theta,\mathcal{D}) +  R(\Theta)$ as the objective.} 
We automatically search for the hyperparameters $\lambda_\theta$, $\lambda_\eta$ and $\lambda_\gamma$ on a validation set using Whetlab~\cite{snoek2012practical}.

\eat{
We define an objective function for training $p(\mathbf{y}_t|\cdot)$ as the variational lower bound of the log-likelihood function. 

\begin{align}
 \log p(\y_{1:t} | \x) & = \log \prod_t p(\y_t| \x,\y_{1:t-1})\\
& =  \sum_t \log \sum_{ s_t\in\{0,1\}} p(\y_t|s_t,\x,\y_{1:t-1} ,s_{1:t-1})
 p(s_t|\x,\y_{1:t-1},s_{1:t-1})\\
& \geq \sum_t \sum_{ s_t\in\{0,1\}} p(s_t|\x,\y_{1:t-1},s_{1:t-1}) \log p(\y_t|s_t,\x,\y_{1:t-1} ,s_{1:t-1}) \\
&= \sum_t \gamma^0_t \log p(\y_t|s_t=0,\x,\y_{-t} ,s_{-t}) + \gamma^1_t \log p(\y_t|s_t=1,\x,\y_{-t} ,s_{-t}).
\end{align}
}

\secmoveup
\section{An Image Caption Dataset with Sentiments}
\label{sec:mturk}

\begin{figure*}[tb]
\centering
\includegraphics[width=.9\textwidth]{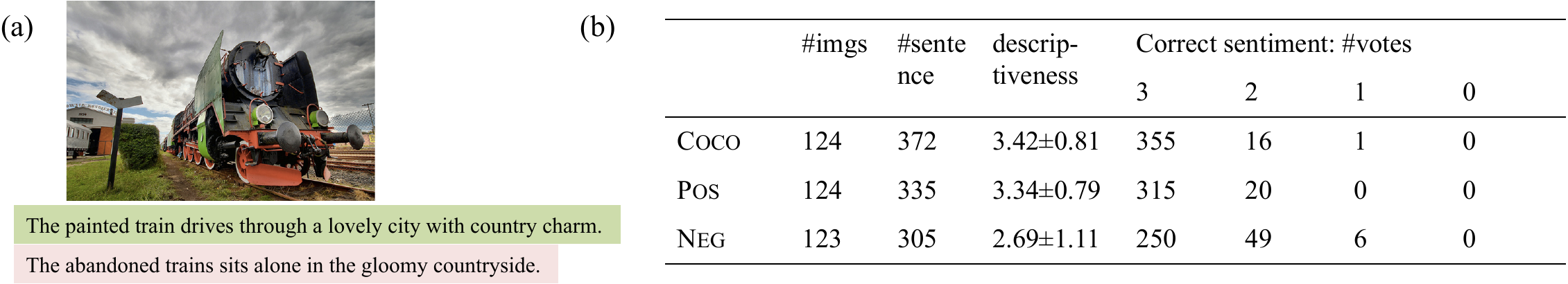}
\caption{(a) One example image with both \colorbox{green!20}{positive} and \colorbox{red!20}{negative} captions written by AMT workers. (b) Summary of quality validation for sentiment captions. The rows are MSCOCO~\protect\shortcite{chen2015microsoft}, and captions with {\sc Pos}itive and {\sc Neg}ative sentiments, respectively. {\em Descriptiveness $\pm$ standard deviation} is rated as 1--4 and averaged across different AMT workers, higher is better. The {\em Correct sentiment} column records the number of captions receiving 3, 2, 1, 0 votes for having a sentiment that matches the image, from three different AMT workers. 
}
 \label{fig:mturk_eval}
\end{figure*}

In order to learn the association between images and captions with sentiments, 
we build a novel dataset of image-caption pairs 
where the caption both describes an image, and also convey the desired sentiment.
\rev{We summarize the new dataset, and the crowd-sourcing task to collect image-sentiment caption data. More details of the data collection process are included in the suplementary\footnotemark[1].}    
\eat{In the rest of this section, we summarize the desired properties and 
components of emotional captions, the design of a crowd-sourcing task to 
collect image-sentiment caption pairs, and crowd-sourced validation of the resulting dataset.}

\eat{In particular, sentiments expressed by the author/owner of the photo, or someone that the owner is communicating to, could be elicited by personal or shared context not contained in the photo itself -- such contextual reasoning is out of scope for automatic captioning from one single image. }

There are many ways a photo could evoke emotions. In this work, we focus on creating a collection and learning sentiments {\em from an objective viewer} who does not know the back story outside of the photo -- a setting also used by recent collections of objectively descriptive image captions~\cite{chen2015microsoft,hodosh2013framing}.  

\bfheading{Dataset construction.} We design a crowd-sourcing task to collect such objectively described emotional image captions. 
This is done in a caption re-writing task based upon objective captions from MSCOCO~\cite{chen2015microsoft} by asking Amazon Mechanical Turk (AMT) workers to choose among ANPs of the desired sentiment, and incorporate one or more of them into any one of the five existing captions. Detailed design of the AMT task is in the appendix\footnotemark[1]\footnotetext[1]{\surl{http://users.cecs.anu.edu.au/~u4534172/senticap.html}}.

The set of candidate ANPs required for this task is collected from the captions for a large sets of online images. We expand the Visual SentiBank~\cite{borth2013sentibank} vocabulary with a set of ANPs from the YFCC100M image captions~\cite{thomee2015yfcc100m} \rev{as} the overlap between the original SentiBank ANPs and the MSCOCO images is insuffcient. We keep ANPs with non-trival frequency and a clear positive or negative sentiment, when rated in the same way as SentiBank. This gives us 1,027 ANPs with a positive emotion, 436 with negative emotions. We collect at least 3 positive and 3 negative captions per image. Figure~\ref{fig:mturk_eval}(a) contains one example image and its respective positive and negative caption written by AMT workers. We release the list of ANPs and the captions in the online appendix\footnotemark[1].

\eat{Appendix~\ref{ssec:cocorewrite} discusses the three design iterations that led to this task, and Figure~\ref{fig:amt_authoring} contains the AMT interface and instructions.}
\eat{Appendix~\ref{ssec:sentibank} describes the construction of the ANP vocabulary in detail.}

\bfheading{Quality validation.} We validate the quality of the resulting captions with another \rev{two-question} AMT task as detailed in the suppliment\footnotemark[1]. This validation is done on \rev{124 images with 3 neutral captions from MSCOCO}, and images with 3 positive and 3 negative captions from our dataset.
We first ask AMT workers to rate the descriptiveness of a caption for a given image on a four-point scale~\cite{hodosh2013framing,vinyals2015show}.  
The {\em descriptiveness} column in Figure~\ref{fig:mturk_eval}(b), shows that the measure for objective descriptiveness tend to decrease when the caption contains additional sentiment. Ratings for the positive captions ({\sc Pos}) have a small decrease (by 0.08, or one-tenth of the standard deviation), while those for the negative captions ({\sc Neg}) have a significant decrease (by 0.73), likely because the notion of negativity is diverse. 

\rev{We also ask whether the sentiment of the sentence matches the image. Each rating task is completed by 3 different AMT workers.} In the {\em correct sentiment} column of Figure~\ref{fig:mturk_eval}(b), we record the number of votes each caption received for bearing a sentiment that matches the image. We can see that the vast majority of the captions are unanimously considered emotionally appropriate ($94\%$, or 315/335 for {\sc Pos}; $82\%$, or 250/305 for {\sc Neg}).
Among the captions with less than unanimous votes received, most of them (20 for {\sc Pos} and 49 for {\sc Neg}) still have majority agreement for having the correct sentiment, which is on par with the level of noise (16 for {\sc Coco} captions).

\secmoveup
\section{Experiments} 
\label{sec:exp}

\begin{table*}[tb]
  \centering \small
    \begin{tabular}{p{.6cm} l | p{.7cm} | p{.53cm}p{.55cm}p{.55cm}p{.55cm} p{.9cm}p{1.0cm}p{.8cm} | p{0.8cm}p{1.5cm}p{1.3cm} }
  \toprule
   & & {\sc sen}\% & {\sc B-1} & {\sc B-2} & {\sc B-3}& {\sc B-4} & {\sc Rouge}$_L$ & {\sc Meteor} & {\sc Cide}$_r$  & {\sc Senti}  & {\sc Desc} & {\sc DescCmp}
   \\ \midrule
  \multirow{3}{*}{\sc Pos} & 
  CNN+RNN & 1.0 & 48.7 & 28.1 & 17.0 & 10.7 & 36.6 & 15.3 & 55.6 &--  & 2.90$\pm$0.90 & --\\
    & ANP-Replace & 90.3 & 48.2 & 27.8 & 16.4 & 10.1 & 36.6 & 16.5 & 55.2  & 84.8\%  & 2.89$\pm$0.92 & 95.0\% \\
    & ANP-Scoring & 90.3 & 48.3 & 27.9 & 16.6 & 10.1 & 36.5 & 16.6 & 55.4  & 84.8\%  & 2.86$\pm$0.96 & 95.3\%\\
    & RNN-Transfer & 86.5 & 49.3 & 29.5 & 17.9 & 10.9 & 37.2 & 17.0 & 54.1 & 84.2\%   & 2.73$\pm$0.96 & 76.2\%\\ 
    & SentiCap & 93.2 & 49.1 & 29.1 & 17.5 & 10.8 & 36.5 & 16.8 & 54.4 & 88.4\%   & 2.86$\pm$0.97 & 84.6\%\\ \midrule
    \multirow{3}{*}{\sc Neg} & 
    CNN+RNN & 0.8  & 47.6 & 27.5 & 16.3 & 9.8 & 36.1 & 15.0 & 54.6  &--  & 2.81$\pm$0.94 & --\\
      & ANP-Replace & 85.5 & 48.1 & 28.8 & 17.7 & 10.9 & 36.3 & 16.0 & 56.5  & 61.4\%  & 2.51$\pm$0.93 & 73.7\%\\
      & ANP-Scoring & 85.5 & 47.9 & 28.7 & 17.7 & 11.1 & 36.2 & 16.0 & 57.1 & 64.5\%   & 2.52$\pm$0.94 & 76.0\%\\
      & RNN-Transfer & 73.4 & 47.8 & 29.0 & 18.7 & 12.1 & 36.7 & 16.2 & 55.9  & 68.1\%  & 2.52$\pm$0.96 & 70.3\%\\ 
      & SentiCap & 97.4 & 50.0 & 31.2 & 20.3 & 13.1 & 37.9 & 16.8 & 61.8  & 72.5\%  & 2.40$\pm$0.89 & 65.0\% \\ \bottomrule
  \end{tabular}
  \caption{Summary of evaluations on captions with sentiment. Columns: {\sc sen}\% is the percentage of output sentences with at least one ANP; {\sc B-1} \ldots {\sc Cider}$_r$ are automatic metrics as described \rvx{in Section~\protect\ref{sec:exp}}; where {\sc B-n} corresponds to the {\sc BLEU-n} metric measuring the co-occurrences of n-grams. {\sc Senti} is the fraction of images for which at least two AMT workers agree that it is the more positive/negative sentence; {\sc Desc} contains the mean and std of the 4-point descriptiveness score, larger is better. {\sc DescCmp} is the percentage of times the method was judged as descriptive or more descriptive than the CNN+RNN baseline. } 
  \label{tab:senticap}\captionmoveup
  \vspace{-5.mm}
\end{table*}

\bfheading{Implementation details.}
We implement RNNs with LSTM units using the Theano package~\cite{BastienTheano2012}. Our implementation of CNN+RNN reproduces caption generation performance in recent work~\cite{Karpathy2015CVPR}.  
The visual input to the switching RNN is 4096-dimensional feature vector from the second last layer of the Oxford VGG CNN~\cite{Simonyan2015}. These features are \rev{linearly embedded into a $D=512$ dimensional space}. Our word embeddings $\Ey$ are 512 dimensions and the hidden state $\h$ and memory cell $\vc$ of the LSTM module also have 512 dimensions. The size of our vocabulary for generating sentences is 8,787, and becomes 8,811 after including additional sentiment words.\eat{from the visual sentiment ontology.} 

\rev{We train the model using Stochastic Gradient Descent (SGD) with mini-batching and the momentum update rule. Mini-batches of size 128 are used with a fixed momentum of 0.99 and a fixed learning rate of 0.001. Gradients are clipped to the range $[-5, 5]$ for all weights during back-propagation. We use perplexity as our stopping criteria. The entire system has about 48 million parameters, and learning them on the sentiment dataset with our implementation takes about 20 minutes at 113 image-sentence pairs per second, while the original model on the MSCOCO dataset takes around 24 hours at 352 image-sentence pairs per second}. Given a new image, we predict the best caption by doing a beam-search with beam-size 5 for the best words at each position. We implementd the system on a multicore workstation with an Nvidia K40 GPU. 

\eat{; which helps to reduce the chances of exploding gradients during the early phases of training.}

\eat{The training procedure stops if perplexity does not improve over ten consecutive check-points (defined as every fifth mini-batch).}
\eat{
Oxford VGG features

D=512

mini-batch of size $128$ sentences
beam search, beam size is 5
gradient clipping on individual weights, gradients clipped to [-5, 5]
}

\bfheading{Dataset setup.} The background RNN is learned on the MSCOCO training set~\cite{chen2015microsoft} of 413K+ sentences on 82K+ images. We construct an additional set of caption with sentiments as described in Section~\ref{sec:mturk} using images from the MSCOCO validation partition. 
The {\sc Pos} subset contains 2,873 positive sentences and 998 images for training, and another 2,019 sentences over 673 images for testing. The {\sc Neg} subset contains 2,468 negative sentences and 997 images for training, and another 1,509 sentences over 503 images for testing.  Each of the test images has three positive and/or three negative captions. 

\bfheading{Systems for comparison.}
The starting point of our model is the RNN with LSTM units and CNN input~\cite{vinyals2015show} learned on the MS COCO training set only, denoted as {\em CNN+RNN}.  \rev{Two simple baselines {\em ANP-Replace} and {\em ANP-Scoring} use sentences generated by {\em CNN+RNN} and then add an adjective with strong sentiment to a random noun. {\em ANP-Replace} adds the most common adjective, in the sentiment captions for the chosen noun. {\em ANP-Scoring} uses multi-class logistic regression to select the most likely adjective for the chosen noun, given the Oxford VGG features.}
The next model, denoted as {\em RNN-Transfer}, learns a fine-tuned RNN \rev{on the sentiment dataset} with additional regularization from {\em CNN+RNN}~\cite{schweikert2008empirical}, \rev{as in $R(\Theta)$ (cf. Eq~\eqref{eq:modreg})}. 
We name the \rev{full switching RNN system as} {\em SentiCap}, which jointly learns \rev{the RNN and the switching probability with word-level sentiments\eat{, i.e. ${\cal L}(\Theta,\mathcal{D}) +  R(\Theta)$} from Equation (\ref{eq:giantr}).\eat{ and (\ref{eq:model1r}).}}

\eat{The next model, denoted as {\em RNN-Transfer}, learns a fine-tuned RNN with additional regularization away from {\em CNN+RNN}~\cite{schweikert2008empirical}, as in Equation (\ref{eq:model1r}). 
We name the combined system {\bf SentiCap}, that jointly learns {\em RNN-Transfer} and the switching probability with word-level reguarilzation, i.e. ${\cal L}(\Theta,\mathcal{D}) +  R(\Theta)$ from Equation (\ref{eq:giantr}) and (\ref{eq:model1r}).}

\begin{figure*}[!th]
  \centering
  \includegraphics[width=.7\textwidth]{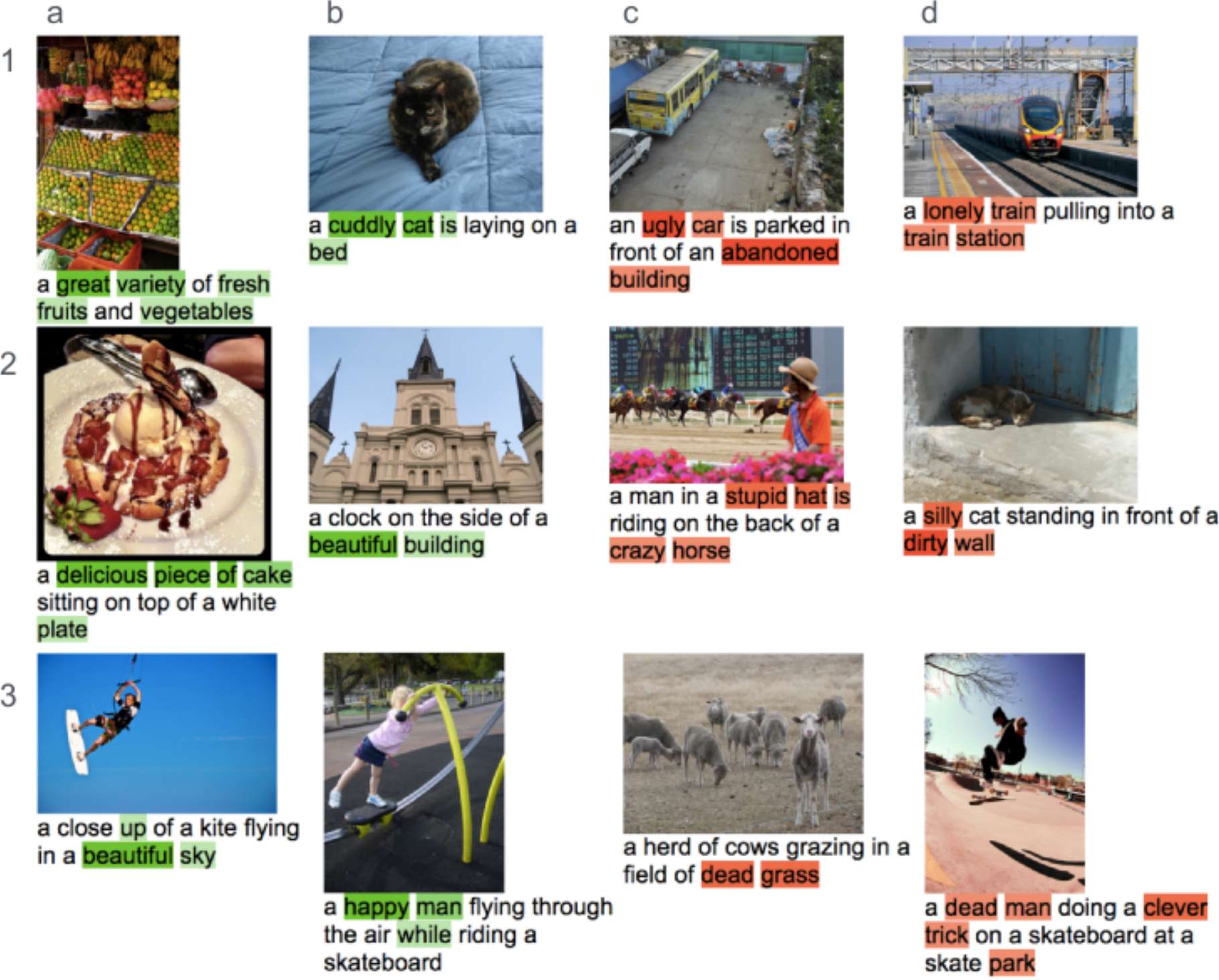}
  \vspace{-3.0mm}
  \secmoveup\caption{Example results from sentiment caption generation. Columns a+b: \textcolor{green!80}{positive} captions; columns c+d: \textcolor{red!50}{negative} captions. Background color indicate the probability of the switching variable $\gamma^1_t=p(s_t|\cdot)$: \colorbox{green}{da}\colorbox{red}{rk} if $\gamma^1_t \ge 0.75$; \colorbox{green!60}{med}\colorbox{red!60}{ium} if $\gamma^1_t \ge 0.5$; \colorbox{green!30}{lig}\colorbox{red!30}{ht} if $\gamma^1_t \ge 0.25$. Row 1 and 2 contain generally successful examples. Row 3 contains examples with various amounts of error in either semantics or sentiment, at times with amusing effects.
  See Section~\ref{sec:exp} for discussions. }
  \label{fig:exp}\captionmoveup
  \vspace{-4.0mm}
\end{figure*}
\bfheading{Evaluation metrics.} We evaluate our system both with automatic metrics and with crowd-sourced judgements through Amazon Mechanical Turk. Automatic evaluation uses the {\sc Bleu}, {\sc Rouge}$_L$, {\sc Meteor}, {\sc Cide}$_r$ metrics from the Microsoft COCO evaluation software~\cite{chen2015microsoft}.

In our crowd-sourced evaluation task AMT workers are given an image and two automatically generated sentences displayed in a random order (example provided in supplement\footnotemark[1]). One sentence is from the {\em CNN+RNN} model without sentiment, while the other sentence is from \rvx{{\em SentiCap} or one of the systems being compared}\eat{either {\em RNN-Transfer} or {\em SentiCap} models trained on captions with sentiment}. AMT workers are asked to rate the descriptiveness of each image from 1-4 and select the more positive or more negative image caption. A process for filtering out noisy ratings is described in the supplement\footnotemark[1]. 
Each pair of sentences is rated by three different AMT workers; at least two must agree that a sentence is more positive/negative for it to be counted as such. The descriptiveness score uses mean aggregation.

\bfheading{Results.} Table \ref{tab:senticap} summarizes the automatic and crowd-sourced evaluations. 
We can see that {\em CNN+RNN} presents almost no sentiment ANPs as it is trained only on MSCOCO. {\em SentiCap} contains significantly more sentences with sentiment words \rev{than \rvx{any of the three}\eat{all three of our} baseline methods,} 
which is expected when the word-level regularization has taken effect. \rev{That {\em SentiCap} has more sentiment words than the two insertion baselines {\em ANP-Replace} and {\em ANP-Scoring} shows that {\em SentiCap} actively drives the flow of the sentence towards using sentimental ANPs. 
Sentences from {\em SentiCap} are, on average, judged by crowd sourced workers to have stronger sentiment than any of the three baselines. For positive {\em SentiCap}, 88.4\% are judged to have a more positive sentiment than the {\em CNN+RNN} baseline. These gains are made with only a small reduction in the descriptiveness \rvx{-- yet this decrease is due to a minority of failure cases, since 84.6\% of captions ranked favorably in the pair-wise descriptiveness comparison.}
{\em SentiCap} negative sentences are judged to have more negative sentiment 72.5\% of the time. On the automatic metrics {\em SentiCap} generating negative captions outperforms all three baselines by a margin.} This improvement is likely due to negative {\em SentiCap} being able to learn more reliable statistics for the new words that only appear in negative ANPs.

{\em SentiCap} sentences with positive sentiment were judged by AMT workers as {\em more interesting} than those without sentiment in 66.4\% of cases, which shows that our method improves the expressiveness of the image captions. On the other hand, negative sentences were judged to be {\em less interesting} than those without sentiment in 63.2\% of cases. This is mostly due to that negativity in the sentence naturally contradicts with being {\em interesting}, a positive sentiment.

It has been noted by \cite{vinyals2015show} that RNN captioning methods tend to exactly reproduce sentences from the training set. Our {\sc SentiCap} method produces a larger fraction of novel sentences than an RNN trained on a single caption domain. A sentence is novel if there is no match in the MSCOCO training set or the sentiment caption dataset. Overall, {\sc SentiCap} produces 95.7\% novel captions; while {\sc CNN+RNN}, which was trained only on MSCOCO, produces 38.2\% novel captions -- higher than the 20\% observed in~\cite{vinyals2015show}.

Figure~\ref{fig:exp} contains a number of examples with generated sentiment captions -- the left half are positive, the right half negative. We can see that the switch variable captures almost all sentiment phrases, and some of the surrounding words (e.g. {\em train station}, {\em plate}). Examples in the first two rows are generally descriptive and accurate such as {\em delicious piece of cake} (2a), {\em ugly car} and {\em abandoned buildings} (1c). Results for the other examples contain more or less inappropriateness in either the content description or sentiment, or both. (3b) captures the {\em happy} spirit correctly, but the semantic of a child in playground is mistaken with that of a man on a skateboard due to very high visual resemblance. \rev{(3d) interestingly juxtaposed the positive ANP {\em clever trick} and negative ANP {\em dead man}, creating an impossible yet amusing caption. }

\secmoveup
\section{Conclusion}
\textmoveup 
\label{sec:conclusion}

We proposed SentiCap, a switching RNN model for generating image captions with sentiments. One novel feature of this model is a specialized word-level supervision scheme to effectively make use of a small amount of training data with sentiments. We also designed a crowd-sourced caption re-writing task to generate sentimental yet descriptive captions. We demonstrate the effectiveness of the proposed model using both automatic and crowd-sourced evaluations, with the SentiCap model able to generate an emotional caption for over 90\% of the images, and the vast majority of the generated captions are rated as having the appropriate sentiment by crowd workers. 
Future work can include unified model for positive and negative sentiment; models for linguistic styles (including sentiments) beyond the word level, and designing generative models for a richer set of emotions such as pride, shame, anger. 


{\small 
\noindent {\bf Acknowledgments}
NICTA is funded by the Australian Government as represented by the
Dept. of Communications and
the ARC through the ICT Centre of Excellence program. 
This work is also supported in part by the Australian Research Council 
via the Discovery Project program. The Tesla K40 used for this research was donated by the NVIDIA Corporation.
}

{\small
\bibliographystyle{aaai}

}

\newpage


\section{Appendix}


\rev{This appendix primarily provides extra details on the model and data collection process. This is included to enusre our results are easily reproducable and to clarify exactly how the data was collected. 

We first provide additional details on the LSTM units used by our approach in Section \ref{sec:app_lstm}. 
Section \ref{ssec:thirdperson} discusses the differences between 1$^{st}$, 2$^{nd}$, and 3$^{rd}$ person sentiment.} See Section \ref{ssec:sentibank} for a discussion of how the ANPs with sentiment where chosen. For details on rewriting sentences to incorporate ANPs see Section \ref{ssec:cocorewrite}. Details on validating the rewritten sentences are in Section \ref{sec:amt_val}. The crowd sourced evaluation of generated sentences is described in Section \ref{sec:app_amt_rating}.

\subsection{The LSTM unit}
\label{sec:app_lstm}

\begin{figure}[ht]
  \centering
  \includegraphics[width=.3\textwidth]{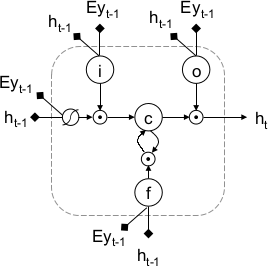}
  \caption{LSTM unit used in our paper, as in Equation~\protect\ref{eq:lstm}. \rev{The filled diamond and square blocks on input nodes represent learn-able weights; in this case parts of the $\mathlarger{\T}^k$ matrix. Note that the weights on these inputs are not the same, they are learned separately.}}
  \label{fig:lstm}
\end{figure}

\rev{The LSTM units we have used are functionally the same as the units used by Vinyals \etal~\shortcite{vinyals2015show}. This differs from the LSTM unit used by Xu \etal~\shortcite{xu2015show} because we do not concatenate contextual information to the units input. A graphical representation of our LSTM units is shown in Figure~\ref{fig:lstm}; for a more complete definition see Equation~\ref{eq:lstm} in the companion paper. In Figure~\ref{fig:lstm}, note that only the LSTM unit is shown, without the fully connected output layers or word embedding layers.}

\subsection{Sentimental descriptions in the first, second, and third person}
\label{ssec:thirdperson}

There are many ways a photo could evoke emotions, they can be referred to as sentiments from the first, second, and third person.  

\begin{figure}[ht]
  \centering
  \includegraphics[width=.35\textwidth]{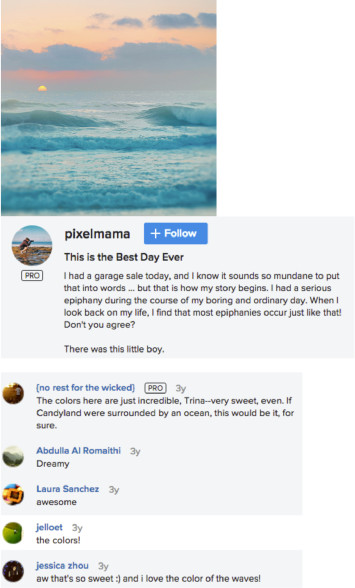}
  \caption{The Flickr photo as discussed in Section~\protect\ref{ssec:thirdperson}. The title and caption are an example of first person sentiment, where a story is told rather than describing the contents of the photo. The comments are second-person sentiments. }
  \label{fig:fp_sent}
\end{figure}

A first person sentiment is for a photo to elicit the emotions of its owner / author / uploader, who then records such sentiment for personal organization or communication to others~\cite{Ames2007whywetag}. Such as the Flickr photo titled ``This is the best day ever''\footnote{\surl{https://www.flickr.com/photos/pixelmama/7612700314/}}, see Figure \ref{fig:fp_sent}. The title and the caption describes a story but not the contents of the photo.

A second person sentiment is expressed by someone whom the photo is {\em communicated to},such as the comments ``awesome'' and ``so sweet'' for the photo above.

The third person sentiment is one expressed by an objective viewer, who has information about its visual content but does not know the backstory, such as describing the photo above as ``Dreamy sunset by the sea''. 

It will be difficult to learn the correct sentiments for the first or second person, since the computer lacks knowledge of the personal and communication context --  to the extent that a change in context and assumptions could completely flip the polarity of the sentiment (See Figure~\ref{fig:mturk_eval}). In this work, we focus on learning possible sentiments from the third person. We collect descriptions with sentiment by people who are asked to describe them -- this setting is close to that of recent collections of subjectively descriptive image captions~\cite{chen2015microsoft,hodosh2013framing}. 

\subsection{Customizing Visual Sentibank for captions}
\label{ssec:sentibank}

Visual SentiBank~\cite{borth2013sentibank} is a database of Adjective-Noun Pairs (ANP) that are frequently used to describe online images. We adopt its methodology to build the sentiment vocabulary. We take the title and the first sentence of the description from the YFCC100M dataset~\cite{thomee2015yfcc100m}, keep entries that are in English, tokenize, and obtain all ANPs that appear in at least 100 images. We score these ANPs using the average of SentiWordNet~\cite{esuli2006sentiwordnet} and SentiStrength~\cite{thelwall2010sentiment}, with the former being able to recognize common lexical variations and the latter designed to score short informal text. We keep ANPs that contain clear positive or negative sentiment, i.e., having an absolute score of 0.1 and above. We then take a union with the Visual SentiBank ANPs. This gives us 1,027 ANPs with a positive emotion, 436 with negative emotions. \rev{A full set of these ANPs are released online, along with sentences containing these ANPs written by AMT workers}.


\subsection{AMT interface for collecting image captions with sentiment}
\label{ssec:cocorewrite}

We went through three design iterations for collecting relevant and succinct captions with the intended sentiment. 

Our first attempt was to invite workers from Amazon Mechanical Turk (AMT) to compose captions with either a positive or negative sentiment for an image -- which resulted in overly long, imaginative captions. 
A typical example is: ``{\em A crappy picture embodies the total cliche of the photographer 'catching himself in the mirror,' while it also includes a too-bright bathroom, with blazing  white walls, dark, unattractive, wood cabinets, lurking beneath a boring sink, holding an amber-colored bowl, that seems completely pointless, below the mirror, with its awkward teenage-composition of a door, showing inside a framed mirror (cheesy, forced perspective,) and a goofy-looking man with a camera.}"

We then asked turkers to place ANPs into an existing caption, which resulted in rigid or linguistically awkward captions. Typical examples include: "a bear that is inside of the great water" and "a bear inside the beautiful water". 

\begin{figure*}[!ht]
  \centering
  \includegraphics[width=.9\textwidth]{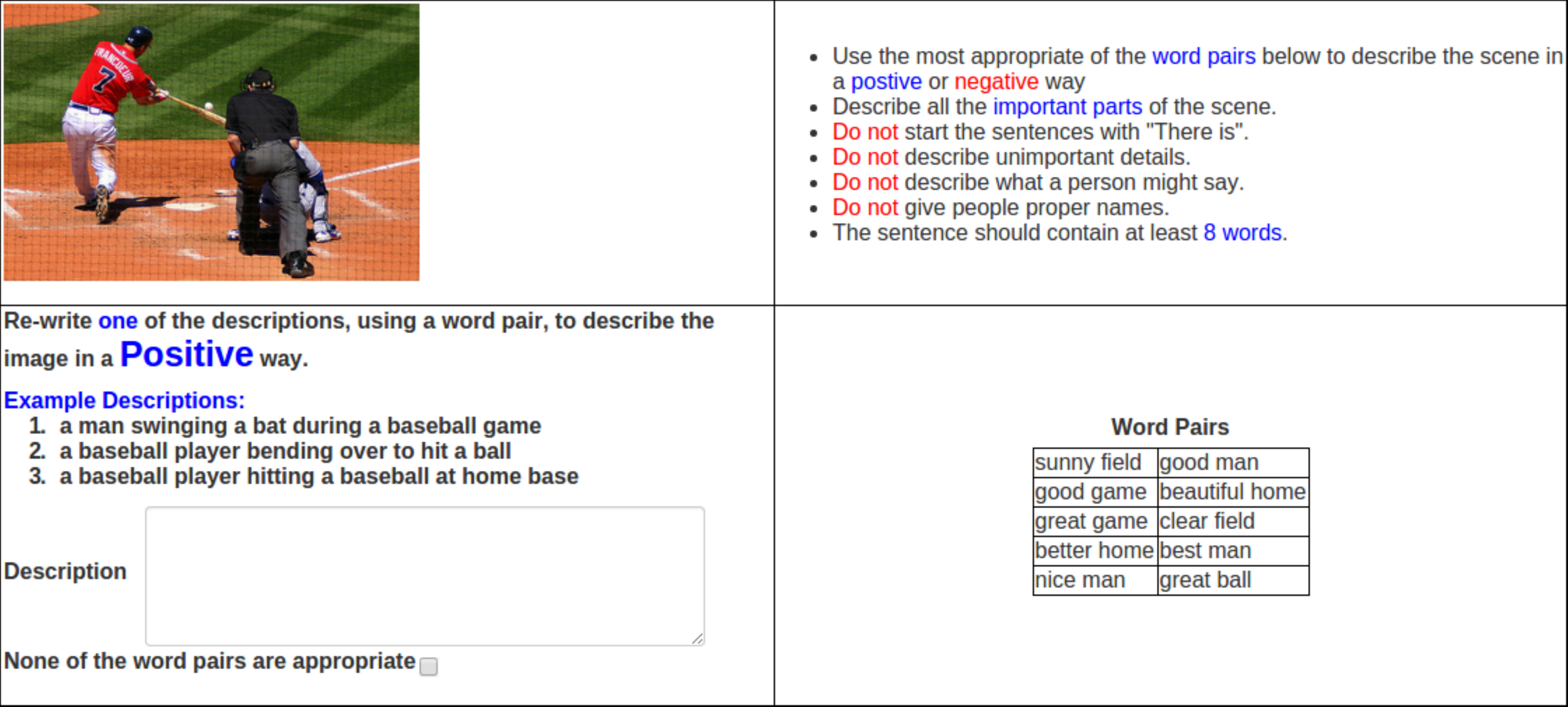}
  \includegraphics[width=.9\textwidth]{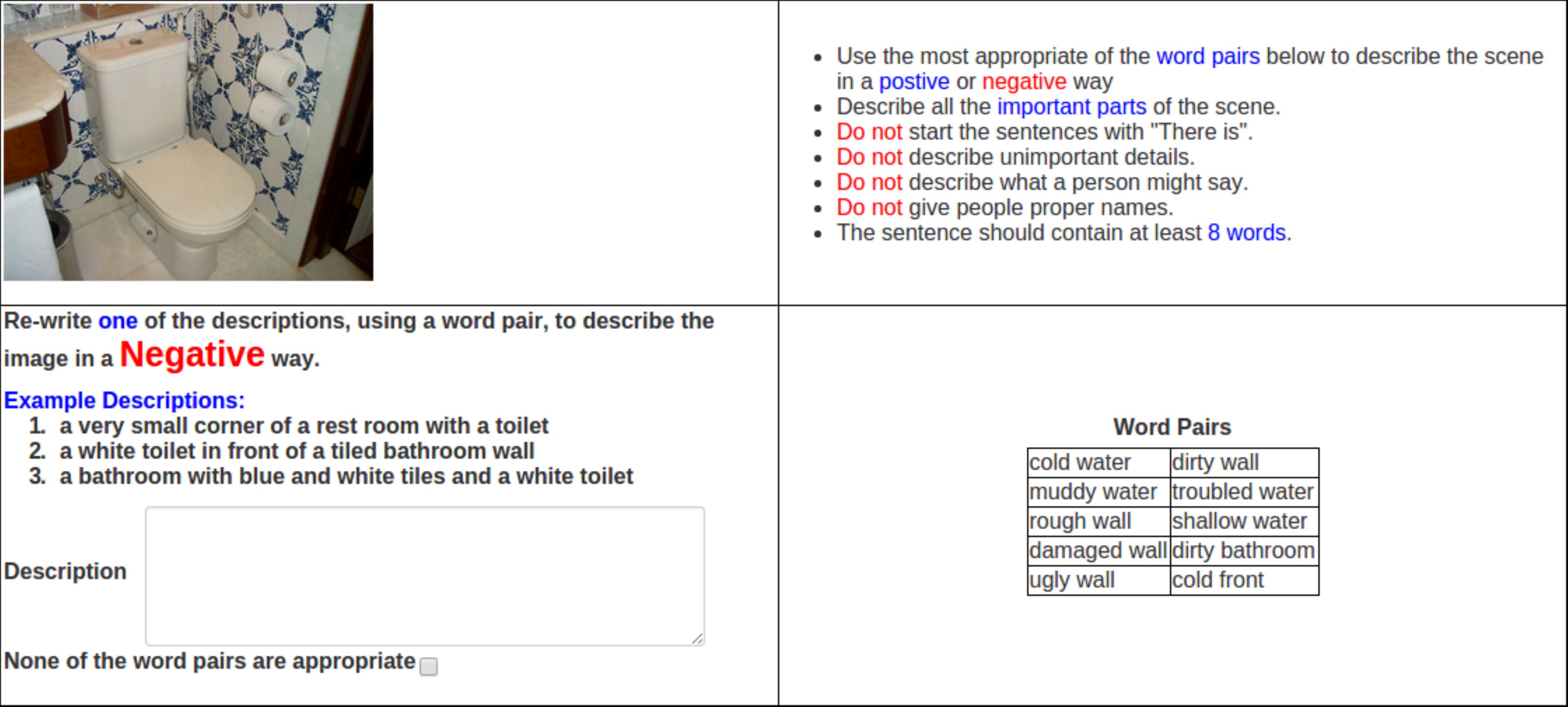}
  \caption{Mturk interfaces and instructions for {\em Collecting} sentences with a positive (top) and negative (bottom) sentiment.}
  \label{fig:amt_authoring}
\end{figure*}

These prompts us to design the following re-writing task: 
we take the available MSCOCO captions, perform tokenization and part-of-speech tagging, and identify nouns and their corresponding candidate ANPs. We provide ten candidate ANPs with the same sentiment polarity and asked AMT worker to rewrite {\em any one of the original captions} about the picture using at least one of the ANPs. The form that the AMT workers are shown is presented in Figure~\ref{fig:amt_authoring}.
We obtained three positive and three negative descriptions for each image, authored by different Turkers. As anecdotal evidence, several turkers emailed to say that this task is {\em very interesting}.

\rev{The instructions given to workers are shown in Figure~\ref{fig:amt_authoring}. We based these instructions on those used by Chen \etal~\shortcite{chen2015microsoft} to construct the MSCOCO dataset. They were modified for brevity and to provide instruction on generating a sentence using the provided ANPs. We found that these instructions were clear to the majority of workers.}

\subsection{AMT interface validating image captions with sentiment}
\label{sec:amt_val}

\begin{figure*}[!ht]
  \centering
  \includegraphics[width=.95\textwidth]{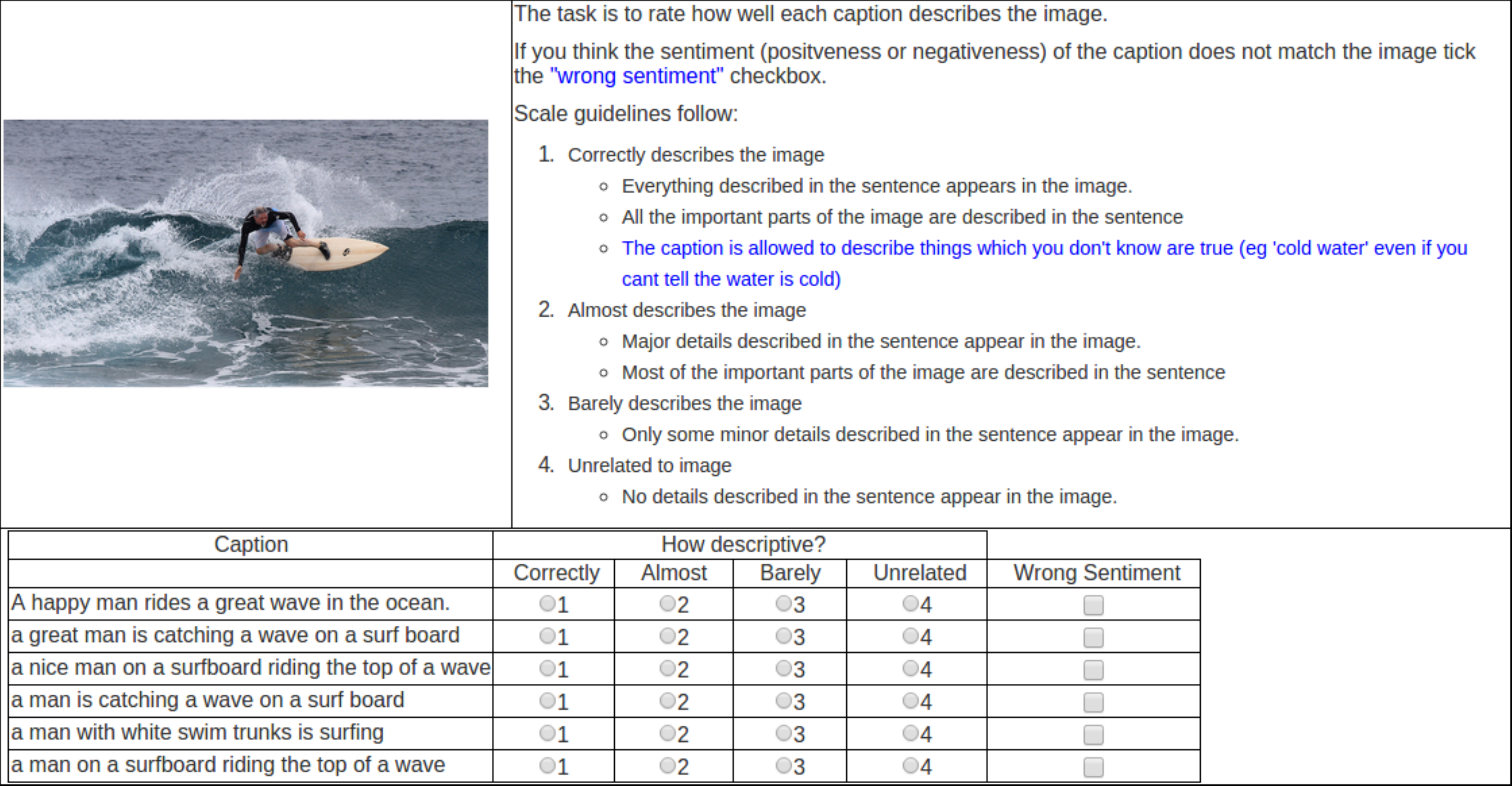}
  \caption{AMT interface and instructions for {\em Rating Groudtruth} sentences}
  \label{fig:amt_rating}
\end{figure*}

\rev{The AMT validation interface, in Figure \ref{fig:amt_rating} was designed to determine what effect adding sentiment into the ground truth captions effects their descriptiveness. Additionally we wanted to understand the fraction of images that could reasonably be described using either positive or negative sentiment. Each task presents the user with three MSCOCO captions and three positive or negative sentences, and asks users to rate them. Our four point descriptiveness scale is based on schemes used by other authors~\cite{hodosh2013framing,vinyals2015show}.}

\subsection{AMT interface for rating captions with a sentiment}
\label{sec:app_amt_rating}

\begin{figure*}[!ht]
  \centering
  \includegraphics[width=.95\textwidth]{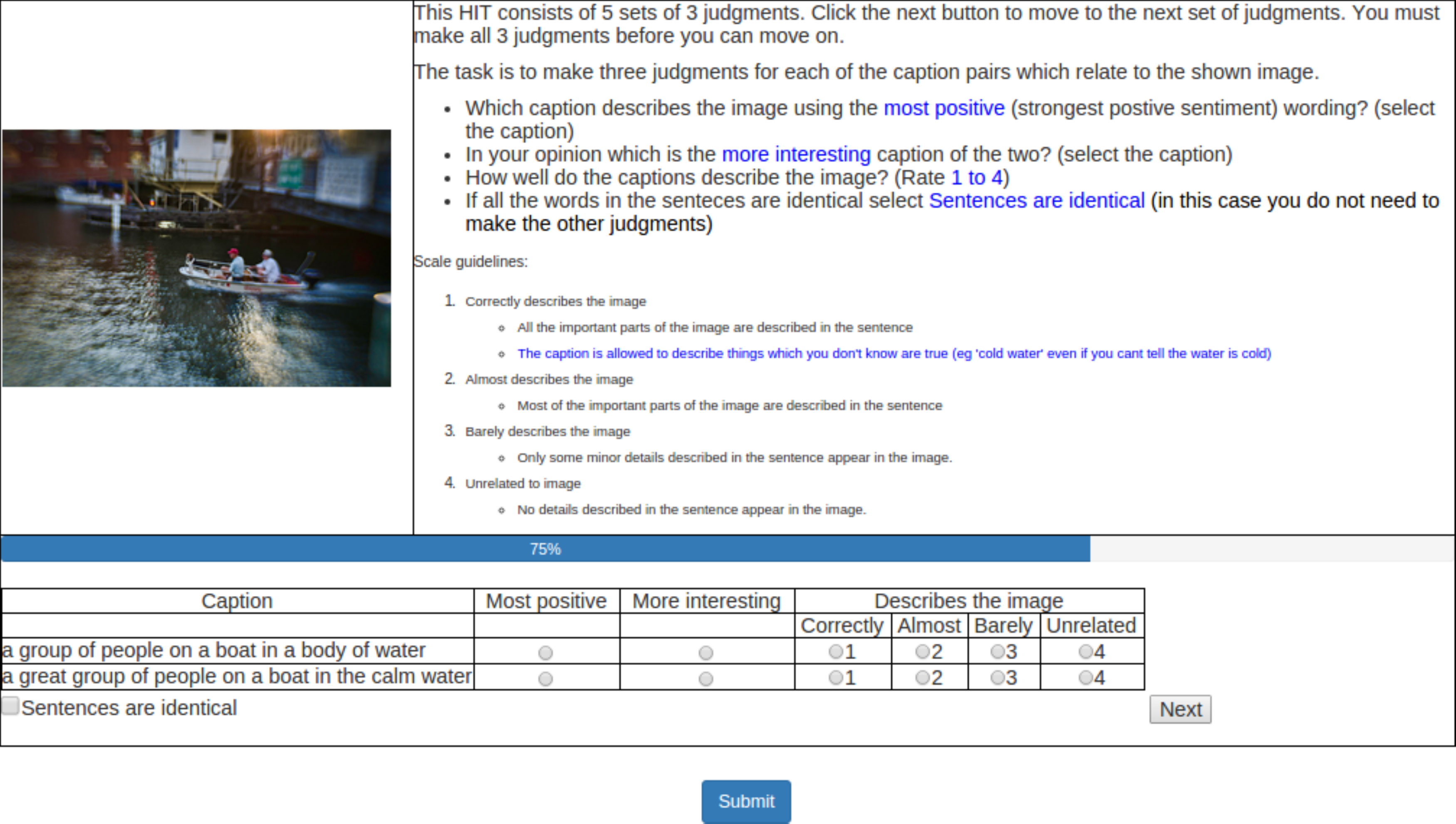}
  \caption{AMT interface and instructions for {\em comparative rating} of generated sentiment sentences}
  \label{fig:amt_eval}
\end{figure*}

\rev{The AMT rating interface shown in Figure \ref{fig:amt_eval} was used to evaluate the performance of the four different methods. Each task consists of three different types of rating: most positive, most interesting and descriptiveness. The most positive and most interesting ratings are done pair-wise, comparing a sentence generated from one of the four methods to a sentence generated by {\em CNN+RNN}. The descriptiveness rating uses the same four point scale as the validation interface from Section \ref{sec:amt_val}. There are 5 images to rate per task; this is essential because of the way AMT calculates prices.

We found that asking Turkers to rate sentences using this method initially produced very poor results, with many Turkers selecting random options without reading the sentences. We suspect that in a number of cases bots were used to complete the tasks. Our first solution was to use more skilled Turkers, called masters workers, although this lead to cleaner results the smaller number of workers meant that a large batch of tasks took far too long to complete. Instead we used workers with a 95\% or greater approval rating. To combat the quality issues we randomly interspersed the manual sentiment captions from our dataset, and then rejected all tasks from worker who failed to achieve 60\% accuracy for the most positive rating. This was found to be an effective way of filtering out the results. We note that there were very few cases where workers were close to the 60\% accuracy cut-off, they were typically much higher or much lower than the threshold, this validates the idea that some workers were not completing the task correctly.}

\end{document}